\documentclass{article}
\usepackage{ijcai19}

\usepackage{times}
\usepackage{soul}
\usepackage{url}
\usepackage[draft]{hyperref}
\usepackage[utf8]{inputenc}
\usepackage[small]{caption}
\usepackage{graphicx}
\usepackage{amsmath}
\usepackage{booktabs}
\usepackage{algorithm}
\usepackage{algorithmic}
\usepackage{epsfig}
\usepackage{amssymb}
\usepackage{mathtools}
\usepackage{multirow}
\usepackage{subcaption}
\usepackage{xcolor}
\usepackage{xspace}
\newcommand{\eat}[1]{}

\DeclarePairedDelimiter{\norm}{\lVert}{\rVert}

\DeclareMathOperator*{\argmin}{arg\,min}

\makeatletter
\DeclareRobustCommand\onedot{\futurelet\@let@token\@onedot}
\def\@onedot{\ifx\@let@token.\else.\null\fi\xspace}

\def\etal{\emph{et al}\onedot}
\makeatother

\title{Beyond Product Quantization: Deep Progressive Quantization for Image Retrieval}

\author{
Lianli Gao$^{1}$\and
Xiaosu Zhu$^{1}$\and
Jingkuan Song$^{1}$\and
Zhou Zhao$^2$\And
Heng Tao Shen$^{1}$\footnote{Contact Author}\\
\affiliations
$^1$Center for Future Media, University of Electronic Science and Technology of China\\
$^2$Zhejiang University\\
\emails
lianli.gao@uestc.edu.cn,
xiaosu.zhu@outlook.com,
jingkuan.song@gmail.com,
zhaozhou@zju.edu.cn,
shenhengtao@hotmail.com
}

\begin{document}

\maketitle

\begin{abstract}
	Product Quantization (PQ) has long been a mainstream for generating an exponentially large codebook at very low memory/time cost. Despite its success, PQ is still tricky for the decomposition of high-dimensional vector space, and the retraining of model is usually unavoidable when the code length changes. In this work, we propose a deep  progressive quantization (DPQ) model, as an alternative to PQ, for large scale image retrieval. DPQ learns the quantization codes sequentially and approximates the original feature space progressively. Therefore, we can train the quantization codes with different code lengths simultaneously. Specifically, we first utilize the label information for guiding the learning of visual features, and then apply several quantization blocks to progressively approach the visual features. Each quantization block is designed to be a layer of a convolutional neural network, and the whole framework can be trained in an end-to-end manner.	Experimental results on the benchmark datasets show that our model significantly outperforms the state-of-the-art for image retrieval. Our model is trained once for different code lengths and therefore requires less computation time. Additional ablation study demonstrates the effect of each component of our proposed model. Our code is released at \url{https://github.com/cfm-uestc/DPQ}.
\end{abstract}
\section{Introduction}
\begin{figure}[t]
	\begin{center}
		\includegraphics[width=1\columnwidth]{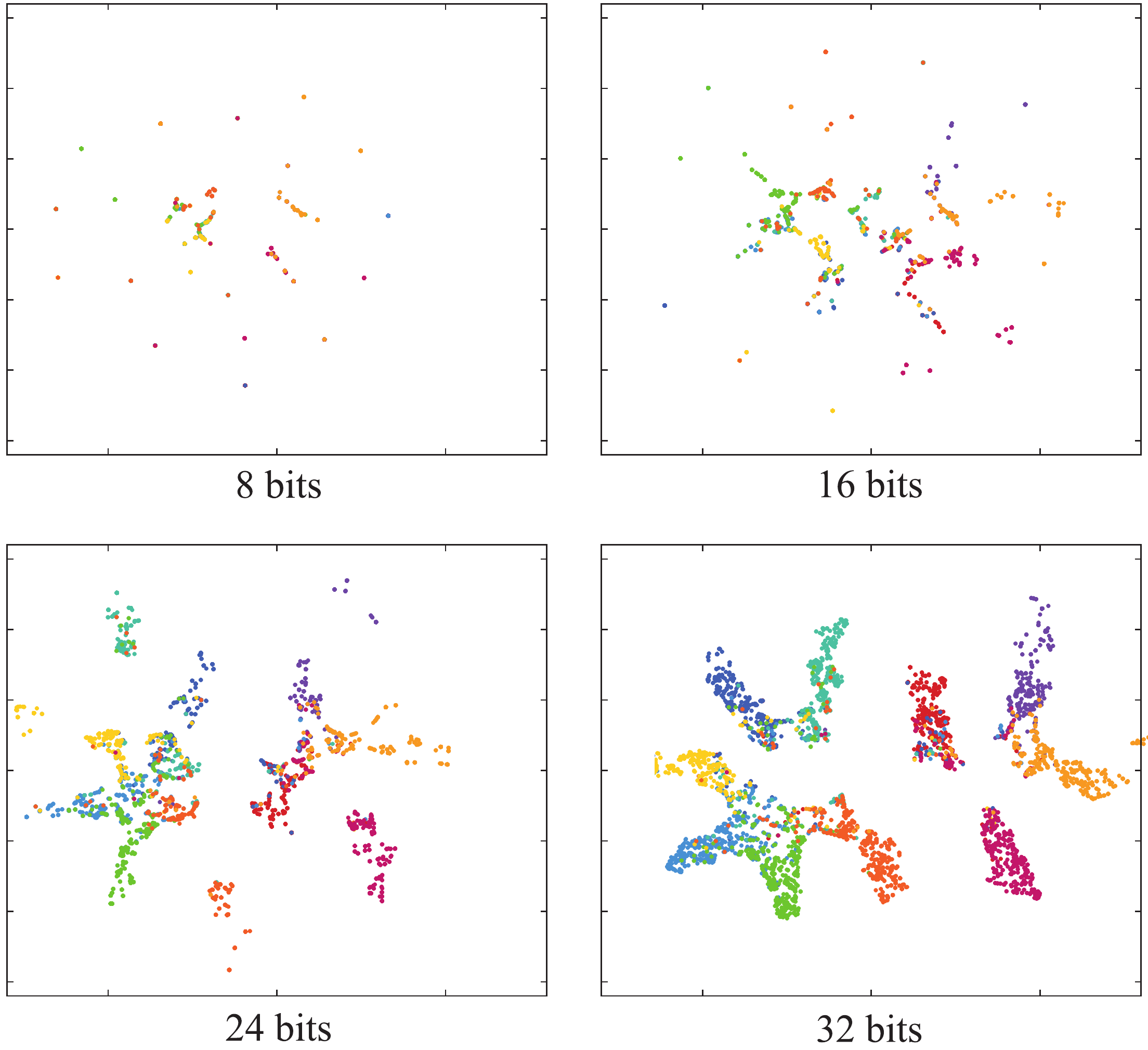}
		\vspace{-0.6cm}
		\caption{t-SNE visualization of progressively generated quantized features (8, 16, 24 and 32 bits) on CIFAR-10. Different colors indicate different labels.}\label{fig.TSNE_All}
	\end{center}
\end{figure}

With the rapidly increasing amount of images, similarity search in large-scale image datasets has been an active research topic in many domains, including computer vision and information retrieval~\cite{HashSurvey}. However, exact nearest-neighbor (NN) search is often intractable because of the size of dataset and the curse of dimensionality for images. Instead, approximate nearest-neighbor (ANN) search is more practical and can achieve orders of magnitude in speed-up compared to exact NN search~\cite{Moraleda08,SongHGXHS18,QBH,VideoH,RetrievalSurvey}.

Recently, hashing methods~\cite{HashSurvey,LSH,KSH,BRE,SDH,DSH,DHN,CNNH,DNNH} are popular for scalable image retrieval due to their compact binary representation and efficient Hamming distance calculation. Such approaches embed data points to compact binary codes by hash functions, while the similarity between vectors are preserved by Hamming distance.

On the other hand, quantization-based methods aim at minimizing the quantization error, and have been shown to achieve superior accuracy~\cite{DVSQ,CQ} over hashing methods, with sacrifice of efficiency.
For input data $\textbf{X} \in \mathbb{R}^{N \times D}$ which contains $N$ samples of $D$ dimensions, Vector Quantization \cite{VQ} tries to create a codebook $\boldsymbol{C} \in \mathbb{R}^{K \times D}$ which contains $K$ codewords $\textbf{c}(k) \in \mathbb{R}^D, k=1,2, \cdots, K$, and assigns each data $\textbf{x}_i, i=1,2, \cdots, N$ to its nearest codeword by a quantizer $q(\textbf{x}_i)= \textbf{c}(e(\textbf{x}_i))$.
In information theory, the function $e(\cdot)$  is called an encoder, and $\textbf{c}(\cdot)$ is called a decoder~\cite{VQ}. The goal of quantization is to minimize the distortion between raw data and quantized data:
\begin{equation}
min \sum\nolimits_i{\norm{\textbf{x}_i - q(\textbf{x}_i)}_2}
\end{equation}
After encoding, a $D$-dimensional data becomes $e(\textbf{x}_i) \in [1,2,...,K]$, which can be represented as a compact binary code $\textbf{b}_i$ of length $\log_2 K$. Product Quantization~\cite{PQ} and Optimized Product Quantization~\cite{OPQ} first divide the feature space $\mathbb{R}^D$ to $M$ subspaces (OPQ also performs a rotation on the data) and perform previous minimization on each subspace. Obviously, the codebooks increase from $1$ to $M$. Composite Quantization~\cite{CQ} also learns $M$ codebooks, but its codewords have the same dimension as the original features. Similar to CQ, Stacked Quantization~\cite{SQ} also uses the sum of multiple codewords to approximate the raw data. But Stacked Quantization uses residual of the quantized results and proposes a hierarchical structure so that the minimization can be performed on each codebook. By integrating deep learning to quantization methods, Cao \etal proposed Deep Quantization Network~\cite{DQN}. It is the first deep learning structure that learns feature by pairwise cosine distance. Then, Cao \etal proposed Deep Visual-Semantic Quantization~\cite{DVSQ} which projects the feature space to semantic space. Inspired by NetVLAD~\cite{NetVLAD}, Benjamin \etal proposed Deep Product Quantization~\cite{DPQ} that enabled differentiable quantization.

To generate an exponentially large codebook at very low memory/time cost, a product quantizer is still the first option. Despite its success, PQ and its variants have several issues. First, to generate binary codes with different code lengths, a retraining is usually unavoidable. Second, it is still tricky for the decomposition of high-dimensional vector space. Different decomposition strategies may result in huge performance differences. To address these issues, we propose a deep progressive quantization (DPQ) model, as an alternative to PQ, for large scale image retrieval. The contributions of DPQ can be summarized as: 1) DPQ is a general framework which can learn codes with different lengths simultaneously, by approximating the original feature space progressively. Different components of DPQ are replaceable and we instantiate it with a few specific designs of label utilization, input and quantization blocks; 2) Each component in the framework is designed to be differentiable, and thus the whole framework can be trained in an end-to-end manner; and 3) Extensive experiments on the benchmark dataset show that DPQ significantly outperforms the state-of-the-art for image retrieval. Additional ablation study demonstrates the effect of each component of our model.

\begin{figure*}[t]
	\begin{center}
		\includegraphics[width=1\linewidth]{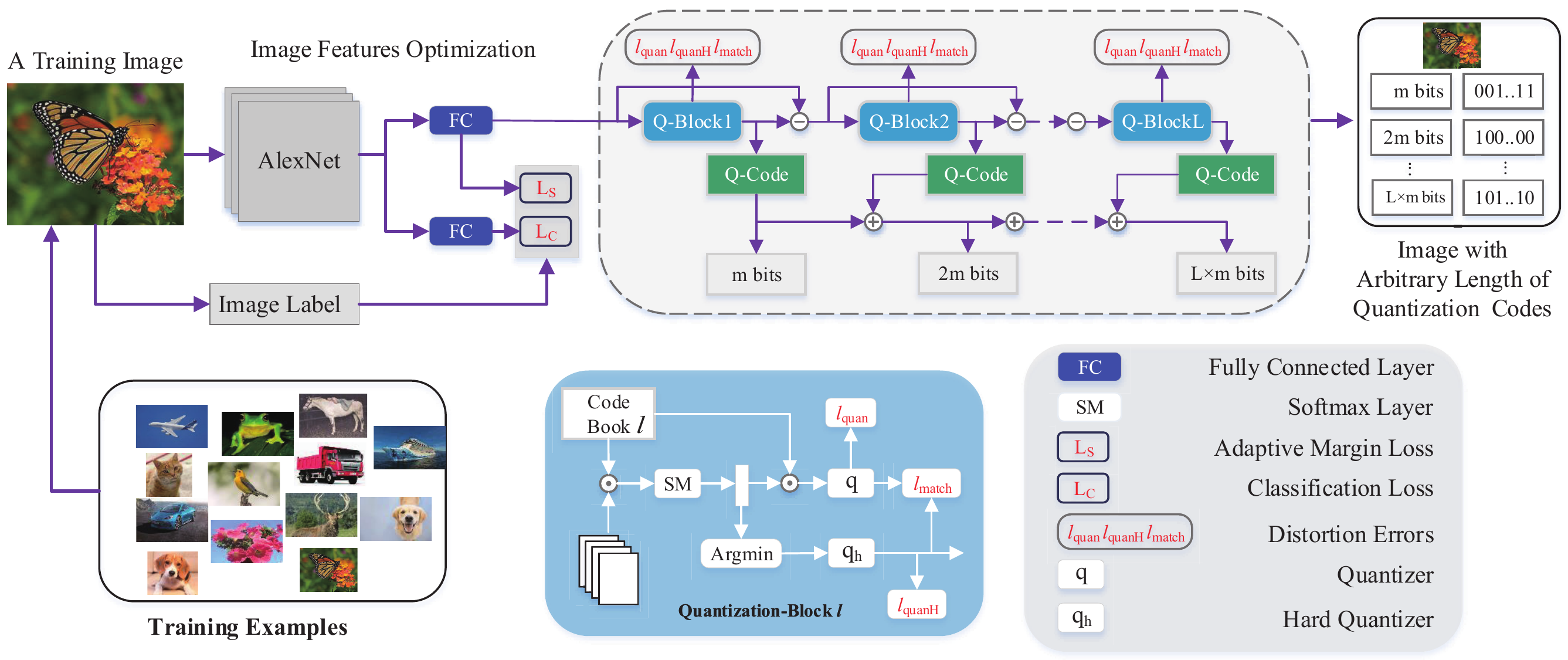}
	\end{center}
	\vspace{-0.5cm}
	\caption{Illustration of our Deep Progressive Quantization for fast image retrieval.}\label{fig.framework}
\end{figure*}

\section{Proposed Method}
Given $N$ images $\textbf{I}$ and a query $\textbf{q}$, the NN search problem aims to find the item $\text{NN}(\textbf{q})$ from $\textbf{I}$ such that its distance to the query is minimum.
We study the ANN search problem and we propose a coding approach called Deep Progressive Quantization (DPQ) for fast image retrieval.

The idea is to approximate a vector $\textbf{x}$ by the composition of several ($L$) elements $\left\{ {\textbf{c}^1({{k_1}}),\textbf{c}^2({{k_2}}),...,\textbf{c}^L({{k_L}})} \right\}$, each of which is selected from a dictionary $\textbf{C}^l$ with $K$ elements, e.g., $\textbf{c}^1({{k_1}})$ is the $k_1$-th element in dictionary $\textbf{C}^1$, and to represent a vector by a short code composed of the indices of selected elements, resulting in a binary codes of length $L \cdot \log_2{K}$ with each element coded by $\log_2{K}$ bits.

We illustrate our proposed DPQ architecture by the scheme in Fig.~\ref{fig.framework}. It consists of two major components: (1) a supervised feature extraction component in a multi-task fashion and (2) a Deep Progressive Quantization component to convert the features to binary codes. In the remainder of this section, we first describe each of them and then illustrate the optimization and search process. 

\subsection{Feature Extraction using Multi-task Learning} \label{MultiTaskProcedure}
For feature extraction, we use AlexNet~\cite{AlexNet}, and we use the output of $fc7$ layer to extract 4096-d features. To fully utilize the supervision information, we introduce two losses based on labels. The first one is a traditional classification loss, which is defined as:
\begin{align}
	L_C(\textbf{x}) &= - \sum_i y_i' \log(y_i) \\
	y_i' &= softmax(\text{MLP}(\textbf{x}))
\end{align}
where $\textbf{x}$ is the input feature, $y_i$ is the label and $y_i'$ is the predicted label. When our images have multi-labels, we modify this loss by multi-label sigmoid cross-entropy:
\begin{align}
	L_C(\textbf{x}) &= - (y'_i) \times y_i + \log{(1 + e^{y'_i})} \\
	y'_i &= \text{MLP}(\textbf{x})
\end{align}

The second one is an adaptive margin loss. We first embed each label into a 300-d semantic vector $\textbf{z}_n$ using word-embedding~\cite{DVSQ}. By applying a fully connected layer on the 4096-d features, we can predict the semantic embedding $\textbf{v}_i$ of each data. Following DVSQ~\cite{DVSQ}, to embed the projects with the semantic label, an adaptive margin loss is defined as:
\begin{align}
	L_S(x)\! &= \!\sum_{i \in \mathcal{Y}_n} \sum_{j \notin \mathcal{Y}_n} (0, \delta_{ij} \!-\! \frac{\boldsymbol{v_i^\intercal}\boldsymbol{z}_n}{\norm{\boldsymbol{v}_i}\norm{\boldsymbol{z}_n}} \!+\! \frac{\boldsymbol{v_j^\intercal}\boldsymbol{z}_n}{\norm{\boldsymbol{v}_j}\norm{\boldsymbol{z}_n}}) \\
	\delta_{ij} &= 1 - \frac{\boldsymbol{v}_i^\intercal \boldsymbol{v}_j}{\norm{\boldsymbol{v}_i}\norm{\boldsymbol{v}_j}}
\end{align}
These two branches can be described as two different tasks.

\subsection{Deep Progressive Quantization} 
\label{DeepStackedQuantizer}
The structure of DPQ is shown in Fig.~\ref{fig.framework}. We denote $\textbf{x}\in \mathbb{R}^D$ as the features of an input image. It goes through $L$ quantization layers (Q-Block), and each Q-Block outputs $m$-bit quantization codes $\textbf{b}^l$ and quantized value $\textbf{q}^l$. The quantization process can be formulated as:
\begin{align}
	\begin{array}{l}
		{\textbf{q}^1} = {Q^1}\left( {{\textbf{x}^1}} \right),
		{\textbf{q}^2} = {Q^2}\left( {{\textbf{x}^2}} \right),
		...,
		{\textbf{q}^L} = {Q^L}\left( {{\textbf{x}^L}} \right)
	\end{array}
	\label{eq.quan}
\end{align}
where ${{\textbf{x}^l}}$ is the input of the $l$-th quantization layer, and it is quantized to ${{\textbf{q}^l}}$ by the $l$-th quantizer $Q^l$. 

The target of DPQ is to progressively approximate $\textbf{x}$ by $\sum\nolimits_{l} {{{\textbf{q}}^i}}$, and objective function can be formulated as:
\begin{equation}
	\left\{\begin{array}{l}
		\ell_\text{quan}^1 = {\textbf{x}} - {\textbf{q}^1} \\
		{\ell _\text{quan}^2} = {\textbf{x}} - \sum\nolimits_{l = 1}^2 {{{\textbf{q}}^l}} \\
		\hspace{1.1cm}\vdots \\
		{\ell _\text{quan}^L} = {\textbf{x}} - \sum\nolimits_{l = 1}^L {{{\textbf{q}}^i}} \\
	\end{array}\right.\\
	\label{eq.quanLoss1}
\end{equation}

The final objective function can be formulated as a weighted sum of quantization losses:
\begin{eqnarray}
	\begin{array}{l}
		\ell_\text{quan} = w^1\ell_\text{quan}^1+w^2\ell_\text{quan}^2+...+w^L\ell_\text{quan}^L\\
	\end{array}
	\label{eq.quanLoss2}
\end{eqnarray}

An important property of this quantization algorithm is that it can generate hash codes with different code lengths simultaneously. In the $l$-th Q-Block, the sum of the $1 \sim l$ quantized value can approximate the original feature $\textbf{x}$, as shown in Eq.~\ref{eq.quanLoss1}. Note that this is a general form for Deep Progressive Quantization. To instantiate DPQ, we need to specify the design of each quantizer $Q^l$ and its input $\textbf{x}^l$.

A straightforward candidate for the Q-Block is K-means algorithm. Given $N$ $D$-dimensional points $\textbf{X}$, the K-means algorithm partitions the database into $K$ clusters, each of which associates one codeword $\textbf{c}_k\in \mathbb{R}^D$. Let $\textbf{C}=[\textbf{c}_1, \textbf{c}_2,...,\textbf{c}_K]\in \mathbb{R}^{D\times K}$ be the corresponding codebook. K-means first randomly initializes a $\textbf{C}$, and then each data point $\textbf{x}_i$ is quantized to:
\begin{equation} 
	\label{NormalQuantization}
	Q_\text{Kmeans}(\textbf{x}_i)\! =\! \argmin_{\textbf{c}(e(\textbf{x}_i))} \norm{\textbf{x}-\textbf{c}(e(\textbf{x}_i))}_2, k=1,\!2,\!\cdots\!,\!K
\end{equation}
Then the codebook is learned by minimizing the within cluster distortion, i.e.,
\begin{equation}
	\mathop {\min }\limits_\textbf{C} \sum\nolimits_i {{{\left\| {{\textbf{x}_i} - Q_\text{Kmeans}({\textbf{x}_i}) } \right\|}_2}}
\end{equation}
This optimization can be done by Lloyd algorithm. However, if we directly use K-means as our Q-Block, it is not an end-to-end model, because $Q_\text{Kmeans}(\textbf{x}_i)$ is \textit{non-differentiable} to $\textbf{c}(i)$. 

To integrate the traditional quantization method into the deep architecture, we have to design a differentiable quantization function. Inspired by NetVLAD~\cite{NetVLAD}, we observe that Eq.~\ref{NormalQuantization} is equivalent to:
\begin{equation}
	Q_\text{Kmeans}(\textbf{x}_i) = \mathop {\lim }\limits_{\gamma  \to  + \infty } \sum\nolimits_k \frac{{{e^{ - \gamma {{\left\| {\textbf{x}_i - {\textbf{c}(k)}} \right\|}_2}}}}}{{\sum\nolimits_k {{e^{ - \gamma {{\left\| {\textbf{x}_i - {\textbf{c}(k)}} \right\|}_2}}}} }} \textbf{c}(k)
\end{equation}
When $\gamma  \to  + \infty$, $\textbf{x}_i$ is quantized to its closest cluster center $\textbf{c}(e(\textbf{x}_i))$. In implementation, choosing a large $\gamma$ can approximate $Q_\text{Kmeans}(\textbf{x})$ well. Therefore, we design our quantization function as:
\begin{equation}
	Q_\text{DPQ}(\textbf{x}) = \sum\nolimits_k \frac{{{e^{ - \gamma d(\textbf{x},\textbf{c}(k))}}}}{{\sum\nolimits_k {{e^{ - \gamma {d(\textbf{x},\textbf{c}(k))}}}} }} \textbf{c}(k)
	\label{eq.quanFun}
\end{equation}
where $d(\textbf{x},\textbf{c}(k))$ indicates the distance between a vector $\textbf{x}$ and a cluster center $\textbf{c}(k)$. Importantly, the proposed quantization function is continuously \textit{differentiable}, which can be readily embedded to any neural networks. $d(\textbf{x},\textbf{c}(k))$ can be any distance function. In practice, we find cosine similarity is better than Euclidean distance, and we define it as:
\begin{equation}
	d(\textbf{x},\textbf{c}(k)) = -\frac{\left\langle {\textbf{x},{\textbf{c}(k)}} \right\rangle}{\norm{\textbf{x}}\norm{\textbf{c}(k)}_2}, i=1,2,\cdots,K
\end{equation}
where $\left\langle{\cdot, \cdot} \right\rangle$ denotes the inner product between two vectors.

We then focus on the input of Q-Block. Following the composition in \cite{SQ} and based on Eq.~\ref{eq.quan}, $\textbf{q}^1$ can be interpreted as a quantizer which approximates the input. But the other quantizers, i.e., $q^l$, can be interpreted as quantizers to approximate the residual of the previous $l$-$1$ quantizers. It is not required that $\textbf{q}^l\approx \textbf{x}^l$. The solution is to let:
\begin{equation}
	\begin{array}{l}
		{\textbf{x}^1} = \textbf{x}\\
		{\textbf{x}^l} = {\textbf{x}^{l - 1}} - {\textbf{q}^{l - 1}}, l > 1
	\end{array}
	\label{eq.input2}
\end{equation}
$q^l$ can be interpreted as a quantizer to approximate the input.

After we determine the quantization functions (Eq.~\ref{eq.quanFun}) and their inputs (Eq.~\ref{eq.input2}), we can calculate the quantization loss based on Eq.~\ref{eq.quan}, Eq.~\ref{eq.quanLoss1} and Eq.~\ref{eq.quanLoss2}. However, the quantization functions (Eq.~\ref{eq.quanFun}) is not a hard assignment, i.e., $Q_\text{DPQ}(\textbf{x})$ is a linear combination of all codewords in codebook. While during the retrieval, each vector is quantized to its closest codeword, named hard assignment, defined as:
\begin{equation}
	Q^H_{\text{DPQ}}(\textbf{x}) = \argmin_{\textbf{c}(e(\textbf{x}))} \norm{\textbf{x} - \textbf{c}(e(\textbf{x}))}_2, k=1,2,\cdots,K
\end{equation}

Therefore, we further define the quantization losses based on the hard assignment as:
\begin{equation}
	\left\{\begin{array}{l}
		\ell_\text{quanH}^1 = {\textbf{x}} - {\textbf{q}_H^1} \\
		{\ell _\text{quanH}^2} = {\textbf{x}} - \sum\nolimits_{l = 1}^2 {{{\textbf{q}}_H^l}} \\
		\hspace{1.1cm}\vdots \\
		{\ell _\text{quanH}^L} = {\textbf{x}} - \sum\nolimits_{l = 1}^L {{{\textbf{q}}_H^L}} \\
	\end{array}\right.\\
\end{equation}
\begin{equation}
	\ell_\text{quanH} = w^1\ell_\text{quanH}^1+w^2\ell_\text{quanH}^2+...+w^L\ell_\text{quanH}^L
	\label{eq.quanHardLoss1}
\end{equation}
where ${\textbf{q}}_H^l$ is the hard quantization value of $\textbf{x}$ in the $l$-th layer.
We further constrain that the learned $\textbf{q}^l$ should be similar to the hard assignment $\textbf{q}_H^l$, formulated as:
\begin{align}
	\ell_\text{match} \!\!=\!\! w^1\norm{\textbf{q}^1\!,\!\textbf{q}_H^1}\!+\!w^2\norm{\textbf{q}^2\!,\!\textbf{q}_H^2}\!+\!...\!+\!w^L\norm{\textbf{q}^L\!,\textbf{q}_H^L}
	\label{eq.matchLoss}
\end{align}
Therefore, the overall distortion loss function is:
\begin{equation}
	E(\boldsymbol{x}) = \ell_\text{quan} + \mu \ell_\text{quanH} + \nu \ell_\text{match}
	\label{DistortionError}
\end{equation}

\subsection{Optimization}

We train our network by optimizing loss functions $(L_S, L_C, E)$. The pipeline of the whole training can be described as follows: Firstly, we collect a batch of training images $\boldsymbol{I}$ as inputs of CNN and extract their features $\boldsymbol{X}$. We then use Q-Block to obtain its quantized features $\boldsymbol{q}_i$, $\boldsymbol{q}_{H,i}$ and associated binary code $\boldsymbol{b}_i$:
\begin{align}
	\boldsymbol{X} = \Theta(\boldsymbol{I}; \theta),~
	\boldsymbol{q}^l = Q_\text{DPQ}^l(\boldsymbol{x}; \pi),
	\boldsymbol{q}_H^l = Q_H^l(\boldsymbol{x}; \pi)\\
	\boldsymbol{b}^l \leftarrow \argmin_{e(\textbf{x})} d({\textbf{x}},{\textbf{c}}(e(\textbf{x}))), e(\textbf{x})=1,2,\cdots,K
\end{align}
where $\theta$ and $\pi$ are the parameters of CNN and Q-Block.

Now, we can use the outputs to calculate total loss function:
\begin{equation} 
	\label{LossFunction}
	L(\boldsymbol{x}; \theta, \pi) = L_S(\boldsymbol{x}) + \lambda L_C(\boldsymbol{x}) + \tau E(\boldsymbol{x})
\end{equation}

Note that the loss function has parameters $\theta$ and $\pi$ to be learned, we use mini-batch gradient descent to update parameters and minimize the loss:
\begin{align}
	\theta \leftarrow \theta - \eta \nabla_{\theta}(L_S, L_C, E),~
	\pi \leftarrow \pi - \eta \nabla_{\pi}(E)
\end{align}
where $\eta$ is the learning rate. We train our network until the maximum iteration is reached.

\subsection{Retrieval Procedure}

After the model is learned, we will obtain a codebook $\pi$. Next, we need to encode the database and perform the search. Given a data point $\textbf{x}_i$ in the database, the $l$-th Q-Block quantizes the input $\textbf{x}_i^l$ to:
\begin{align} 
  &e(\textbf{x}_i^l) = \argmin_{k^l} \norm{\textbf{x}_i^l - \textbf{c}^l({k^l})}^2_2, k^l=1,2,\cdots,K\\
  &Q^H_{\text{DPQ}}(\textbf{x}_i^l) = \textbf{c}^l(e(\textbf{x}_i^l))
\end{align}
And therefore, $\textbf{x}_i$ is quantized by $\overline {\textbf{x}}_i$ and represented by $\textbf{b}_i$:
\begin{align} 
	&\overline {\textbf{x}}_i = \sum\nolimits_l {Q^H_{\text{DPQ}}(\textbf{x}_i^l)} = \sum\nolimits_l {\textbf{c}^l(e(\textbf{x}_i^l))}\\
	&\textbf{b}_i \!=\! [\text{binary}(e(\textbf{x}_i^1)), \text{binary}(e(\textbf{x}_i^2))\!,\!...\!,\!\text{binary}(e(\textbf{x}_1^L))]
\end{align}
where $\text{binary}(.)$ operation is to convert an index in the range of $[1,K]$ to a binary code with the code length of $M$=$\log_2{K}$. Therefore, we can simultaneously obtain $L$ binary codes with different lengths. 

The Asymmetry Quantization Distance (AQD) of a query $\textbf{q}$ to database point $\textbf{x}_i$ is approximated by $\left\| {\textbf{q} - \overline {\textbf{x}} } \right\|_2^2 = \left\| {\textbf{q} - \sum\nolimits_l {\textbf{c}^l(e(\textbf{x}_i^l))} } \right\|_2^2$, which can be reformulated as:
\begin{align}
&\| {q -\! \sum\limits_l {\textbf{c}^l(e(\textbf{x}_i^l))} } \|_2^2 = \sum\limits_l {\| {\textbf{q} - {\textbf{c}^l(e(\textbf{x}_i^l))}} \|} _2^2 \\\nonumber
&- \!( {L \!-\! 1} )\| \textbf{q} \|_2^2 
\!+\!\! \sum\limits_{l1 = 1}^L {\sum\limits_{l2 = 1,l2 \ne l1}^L\!\! {{{\left( { {\textbf{c}^{l1}(e(\textbf{x}_i^{l1}))}} \right)}^T} {\textbf{c}^{l2}(e(\textbf{x}_i^{l2}))}} }
\end{align}
We can see that the first and third term can be precomputed and stored, while the second term is constant. Therefore, the AQD can be efficiently calculated.

\section{Experiments}
\begin{table*}[t]
	\small	
	\centering
	\resizebox{0.9\textwidth}{!}{%
		\begin{tabular}{c|cccc|cccc|cccc}
			\hline
			\multirow{2}{*}{Method} & \multicolumn{4}{c|}{CIFAR-10}        & \multicolumn{4}{c|}{NUS-WIDE}        & \multicolumn{4}{c}{ImageNet}        \\ \cline{2-13} 
			& 8 bits & 16 bits & 24 bits & 32 bits & 8 bits & 16 bits & 24 bits & 32 bits & 8 bits & 16 bits & 24 bits & 32 bits \\ \hline
			ITQ-CCA~\cite{ITQ}       & 0.315  & 0.354   & 0.371   & 0.414   & 0.526  & 0.575   & 0.572   & 0.594   & 0.189  & 0.270   & 0.339   & 0.436   \\
			BRE~\cite{BRE}           & 0.306  & 0.370   & 0.428   & 0.438   & 0.550  & 0.607   & 0.605   & 0.608   & 0.251  & 0.363   & 0.404   & 0.453   \\
			KSH~\cite{KSH}           & 0.489  & 0.524   & 0.534   & 0.558   & 0.618  & 0.651   & 0.672   & 0.682   & 0.228  & 0.398   & 0.499   & 0.547   \\
			SDH~\cite{SDH}           & 0.356  & 0.461   & 0.496   & 0.520   & 0.645  & 0.688   & 0.704   & 0.711   & 0.385  & 0.516   & 0.570   & 0.605   \\
			SQ~\cite{SQ}             & 0.567  & 0.583   & 0.602   & 0.615   & 0.653  & 0.691   & 0.698   & 0.716   & 0.465  & \underline{0.536}   & \underline{0.592}   & \underline{0.611}   \\ \hline
			CNNH~\cite{CNNH}         & 0.461  & 0.476   & 0.465   & 0.472   & 0.586  & 0.609   & 0.628   & 0.635   & 0.317  & 0.402   & 0.453   & 0.476   \\
			DNNH~\cite{DNNH}         & 0.525  & 0.559   & 0.566   & 0.558   & 0.638  & 0.652   & 0.667   & 0.687   & 0.347  & 0.416   & 0.497   & 0.525   \\
			DHN~\cite{DHN}           & 0.512  & 0.568   & 0.594   & 0.603   & 0.668  & 0.702   & 0.713   & 0.716   & 0.358  & 0.426   & 0.531   & 0.556   \\
			DSH~\cite{DSH}           & 0.592  & 0.625   & 0.651   & 0.659   & 0.653  & 0.688   & 0.695   & 0.699   & 0.332  & 0.398   & 0.487   & 0.537   \\
			DQN~\cite{DQN}           & 0.527  & 0.551   & 0.558   & 0.564   & 0.721  & 0.735   & 0.747   & 0.752   & 0.488  & 0.552   & 0.598   & 0.625   \\
			DVSQ~\cite{DVSQ}         & \underline{0.715} & \underline{0.727} & \underline{0.730} & \underline{0.733} & \underline{0.780} & \underline{0.790} & \underline{0.792} & \underline{0.797} & \underline{0.500} & {0.502} & {0.505} & {0.518}   \\ \hline
			DPQ            & \textbf{0.814} & \textbf{0.833} & \textbf{0.834} & \textbf{0.831} & \textbf{0.786} & \textbf{0.821} & \textbf{0.832} & \textbf{0.834} & \textbf{0.521} & \textbf{0.602} & \textbf{0.613} & \textbf{0.623} \\ \hline
		\end{tabular}
	}
	\caption{Quantitative comparison with state-of-the-art methods on on three datasets. The scores reported are mean Average Precision values.}
	\label{tab.Result}
\end{table*}

\begin{figure*}[t]
	\begin{center}
		\includegraphics[width=0.82\paperwidth]{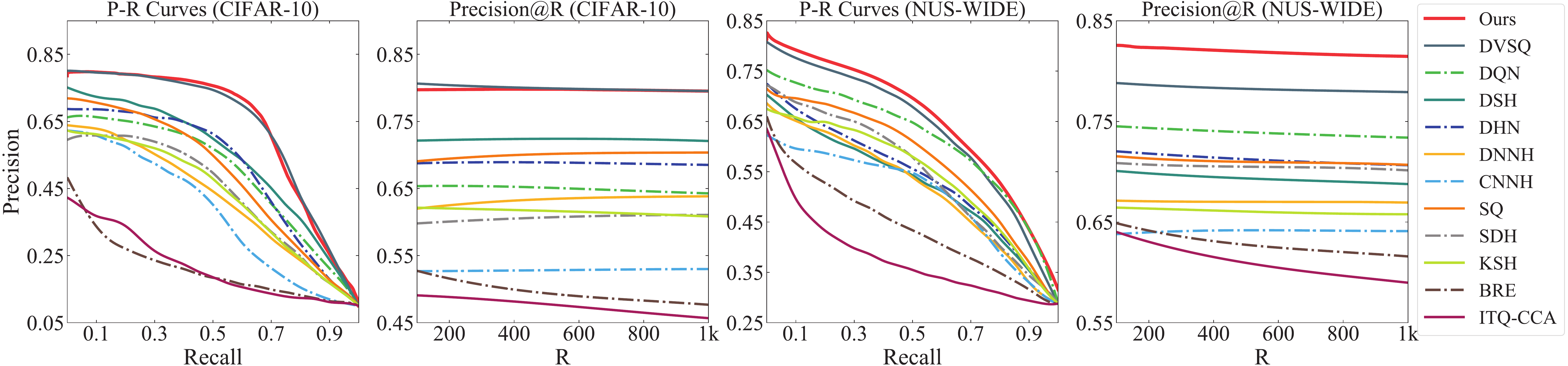}
		\vspace{-0.6cm}
		\caption{Quantitative comparison with state-of-the-art methods on two datasets: CIFAR-10 and NUS-WIDE. For each data set, we demonstrate Precision-Recall (P-R) curves and Precision@R curves. All results are based on 32-bit.}\label{fig.CurveOnFirst}
	\end{center}
\end{figure*}

We evaluate our DPQ on the task of image retrieval. Specifically, we first evaluate the effectiveness of DPQ against the state-of-the-art methods on three datasets, and then conduct ablation study to explore the effect of important components.

\subsection{Setup} 
\label{Setup}
We conduct the experiments on three public benchmark datasets: \textbf{CIFAR-10}, 
\textbf{NUS-WIDE} and \textbf{ImageNet}. 

\textbf{CIFAR-10}~\cite{CIFAR} is a public dataset labeled in 10 classes. It consists of 50,000 images for training and 10,000 images for validation. We follow~\cite{DQN,DVSQ} to combine all images together. Randomly select 500 images per class as the training set, and 100 images per class as the query set. The remaining images are used as database. 

\textbf{NUS-WIDE}~\cite{NUS} consists of 81 concepts, and each image is annotated with one or more concepts. We follow~\cite{DQN,DVSQ} to use the subset of the 21 most frequent concepts (195,834 images). We randomly sample 5,000 images as the query set, and use the remaining images as database. Furthermore, we randomly select 10,000 images from the database as the training set. 

\textbf{ImageNet}~\cite{ImageNet} contains 1.2M images labeled with 1,000 classes. We follow~\cite{DVSQ} to randomly choose 100 classes. We use all images of these classes in the training set and validation set as the database and queries respectively. We randomly select 100 images for each class in the database for training.

We compare our method with state-of-the-art supervised hashing and quantization methods, including 5 shallow-based methods (\textbf{KSH}~\cite{KSH}, \textbf{SDH}~\cite{SDH}, \textbf{BRE}~\cite{BRE}, \textbf{ITQ-CCA}~\cite{ITQ} and \textbf{SQ}~\cite{SQ}) and 6 deep-based methods (\textbf{DSH}~\cite{DSH}, \textbf{DHN}~\cite{DHN}, \textbf{DNNH}~\cite{DNNH}, \textbf{CNNH}~\cite{CNNH}, \textbf{DQN}~\cite{DQN} and \textbf{DVSQ}~\cite{DVSQ}). For shallow-based methods, we use $fc7$ of pre-trained AlexNet as features. We evaluate all methods with 8, 16, 24 and 32 bits.

To evaluate retrieval effectiveness on these datasets, we follow~\cite{DQN,DVSQ,DNNH} to use three evaluation metrics: mean Average Precision (mAP), Precision-Recall curves, and Precision@R (number of returned samples) curves. We follow the settings in~\cite{DQN,DVSQ} to measure mAP@54000 on CIFAR-10, mAP@5000 on NUS-WIDE and mAP@5000 on ImageNet. Following~\cite{DHN,DSH,DQN,DVSQ}, the evaluation is based on Asymmetric Quantization Distance (AQD). 
We use 300-d features as the input of the Q-Block.
In order to compare with other methods using the same bit-length, we construct the codebook with $L = 4$, $K$=$256$=$2^8$, so that each codebook can provide a piece of 8-bit binary code. We set epoch to 64 and batch to 16. We use the Adam optimizer with default value. We tune the learning rate $\eta$ from $10^{-4}$ to $10^{-1}$. As for $\lambda, \tau, \mu, \nu$ in loss function Eq.~\ref{LossFunction}, we empirically set them as $\lambda=0.1, \tau=1, \mu=1, \nu=0.1$. Our implementation is based on Tensorflow.

\begin{table*}[t]
	\small
	\centering
\resizebox{0.85\textwidth}{!}{%
		\begin{tabular}{c|cccc|cccc|cccc}
			\hline
			\multirow{2}{*}{Method} & \multicolumn{4}{c|}{CIFAR-10}        & \multicolumn{4}{c|}{NUS-WIDE}        & \multicolumn{4}{c}{ImageNet}        \\ \cline{2-13} 
			& 8 bits & 16 bits & 24 bits & 32 bits & 8 bits & 16 bits & 24 bits & 32 bits & 8 bits & 16 bits & 24 bits & 32 bits \\ \hline
			$E$                      & 0.172  & 0.188   & 0.184   & 0.205   & 0.472  & 0.620   & 0.684   & 0.713   & 0.248  & 0.265   & 0.277   & 0.283   \\
			%$L_S + L_C$              & 0.365  & 0.337   & 0.324   & 0.318   & 0.682  & 0.718   & 0.722   & 0.722   & 0.057  & 0.088   & 0.111   & 0.136   \\
			$E + L_S$                & \underline{0.741}  & 0.750   & 0.755   & \underline{0.764} & \underline{0.780} & \underline{0.815} & \underline{0.823} & \underline{0.823} & \underline{0.514} & \underline{0.534} & \underline{0.540} & \underline{0.543} \\
			$E + L_C$                & 0.614  & 0.644   & 0.654   & 0.671   & 0.475  & 0.580   & 0.616   & 0.661   & 0.328  & 0.393   & 0.408   & 0.418   \\
			Two-Step                 & {0.732} & \underline{0.755} & \underline{0.758} & 0.760   & 0.505  & 0.626   & 0.656   & 0.700   & 0.345  & 0.399   & 0.419   & 0.412   \\
			No-Soft                  & 0.556  & 0.575   & 0.598   & 0.621   & 0.542  & 0.684   & 0.701   & 0.692   & 0.302  & 0.332   & 0.320   & 0.328   \\ 
\hline
			DPQ                      & \textbf{0.814} & \textbf{0.833} & \textbf{0.834} & \textbf{0.831} & \textbf{0.786} & \textbf{0.821} & \textbf{0.832} & \textbf{0.834} & \textbf{0.521} & \textbf{0.602} & \textbf{0.613} & \textbf{0.623} \\ \hline
		\end{tabular}
	}
	\caption{Ablation Studies: mean Average Precision scores of 6 variants of DPQ on three datasets by setting the quantization code length as 8 bits, 16 bits, 24 bits, 32 bits. Variants are described in sec.~\ref{AblationStudy}.}
	\label{tab.AblationStudy}
\vspace{-0.2cm}
\end{table*}

\begin{figure}[t]
	\begin{center}
		\includegraphics[width=1\columnwidth]{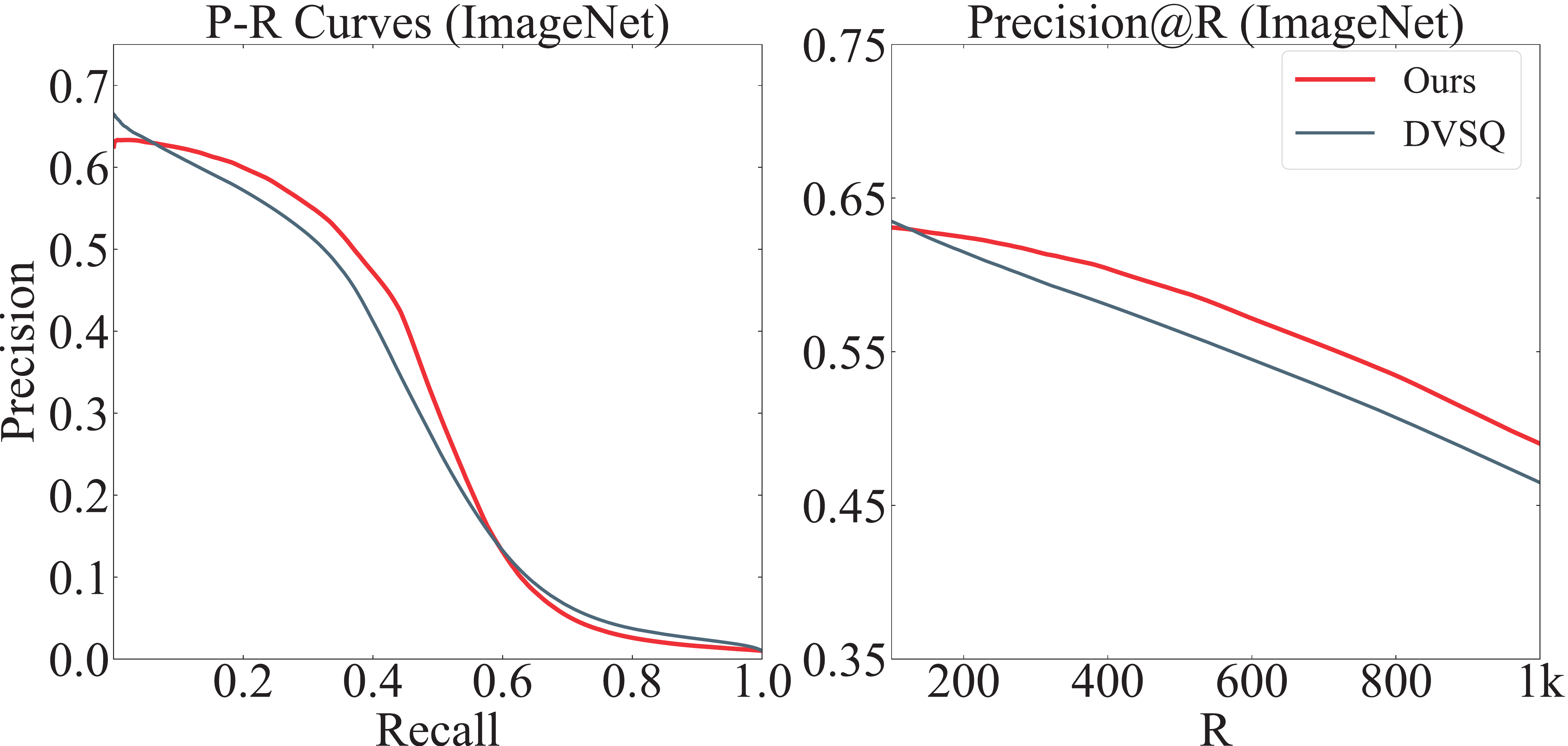}
		\vspace{-0.6cm}
		\caption{Quantitative comparison with DVSQ method on ImageNet. We demonstrate Precision-Recall (P-R) curves and Precision@R curves. All results are based on 32-bit.}\label{fig.CurveOnImagenet}
	\end{center}
\end{figure}

\subsection{Comparison With the State-of-the-Art Methods}
The mAP of compared methods are based on DVSQ~\cite{DVSQ}, and the results are shown in Tab.~\ref{tab.Result}. It can be observed that: 1) Our method (DPQ) significantly outperforms the other deep and non-deep hashing methods in all datasets. In CIFAR-10, the improvement of DPQ over the other methods is more significant, compared with that in NUS-WIDE and ImageNet datasets. Specifically, it outperforms the best counterpart (DVSQ) by 9.9\%, 10.6\%, 10.4\% and 9.8\% for 8, 16, 24 and 32-bits hash codes. DPQ improves the state-of-the-art by 0.6\%, 3.1\%, 4.0\% and 3.7\% in NUS-WIDE dataset, and 2.1\%, 6.6\%, 2.1\%, 1.2\% in ImageNet dataset.
2) With the increase of code length, the performance of most indexing methods is improved accordingly. For our DPQ, the mAP increased by 2.0\%, 4.8\% and 10.2\% for CIFAR-10, NUS-WIDE and ImageNet dataset respectively. The improvement for CIFAR-10 dataset is relatively small. One possible reason is that CIFAR-10 contains simple images, and short codes are good enough to conduct accurate retrieval. 
3) Deep-based quantization methods perform better than shallow-based methods in general. This indicates that jointly learning procedure can usually obtain better features. On the other hand, the performances of two deep hashing methods (CNNH~\cite{CNNH} and DNNH~\cite{DNNH}) are unsatisfactory. A possible reason is that the deep hashing methods use only a few fully connected layers to extract the features, which is not very powerful.

The precision-recall curves and precision@R curves for CIFAR-10 and NUS-WIDE datasets are shown in Fig.~\ref{fig.CurveOnFirst}, and the results for ImageNet dataset are shown in Fig.~\ref{fig.CurveOnImagenet}. In general, curves for different methods are consistent with their performances of mAP.

\begin{figure}[t]
	\begin{center}
		\begin{subfigure}{.126\paperwidth}
			\centering
			\includegraphics[width=1\linewidth,height=0.8\linewidth]{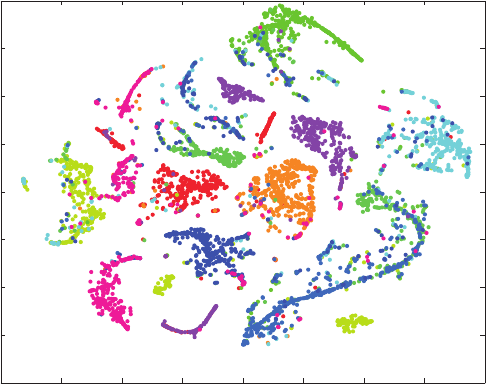}
			\caption{DQN}
			\vspace{-0.3cm}
		\end{subfigure}
		\begin{subfigure}{.126\paperwidth}
			\centering
			\includegraphics[width=1\linewidth,height=0.8\linewidth]{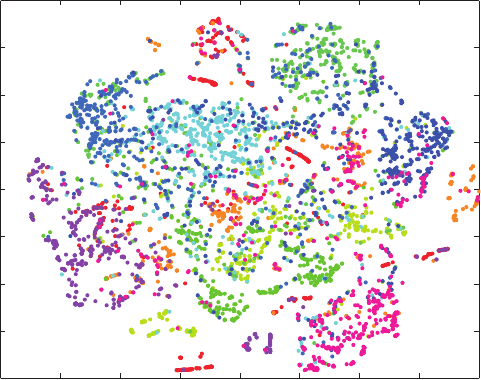}
			\caption{DVSQ}
			\vspace{-0.3cm}
		\end{subfigure}
		\begin{subfigure}{.126\paperwidth}
			\centering
			\includegraphics[width=1\linewidth,height=0.8\linewidth]{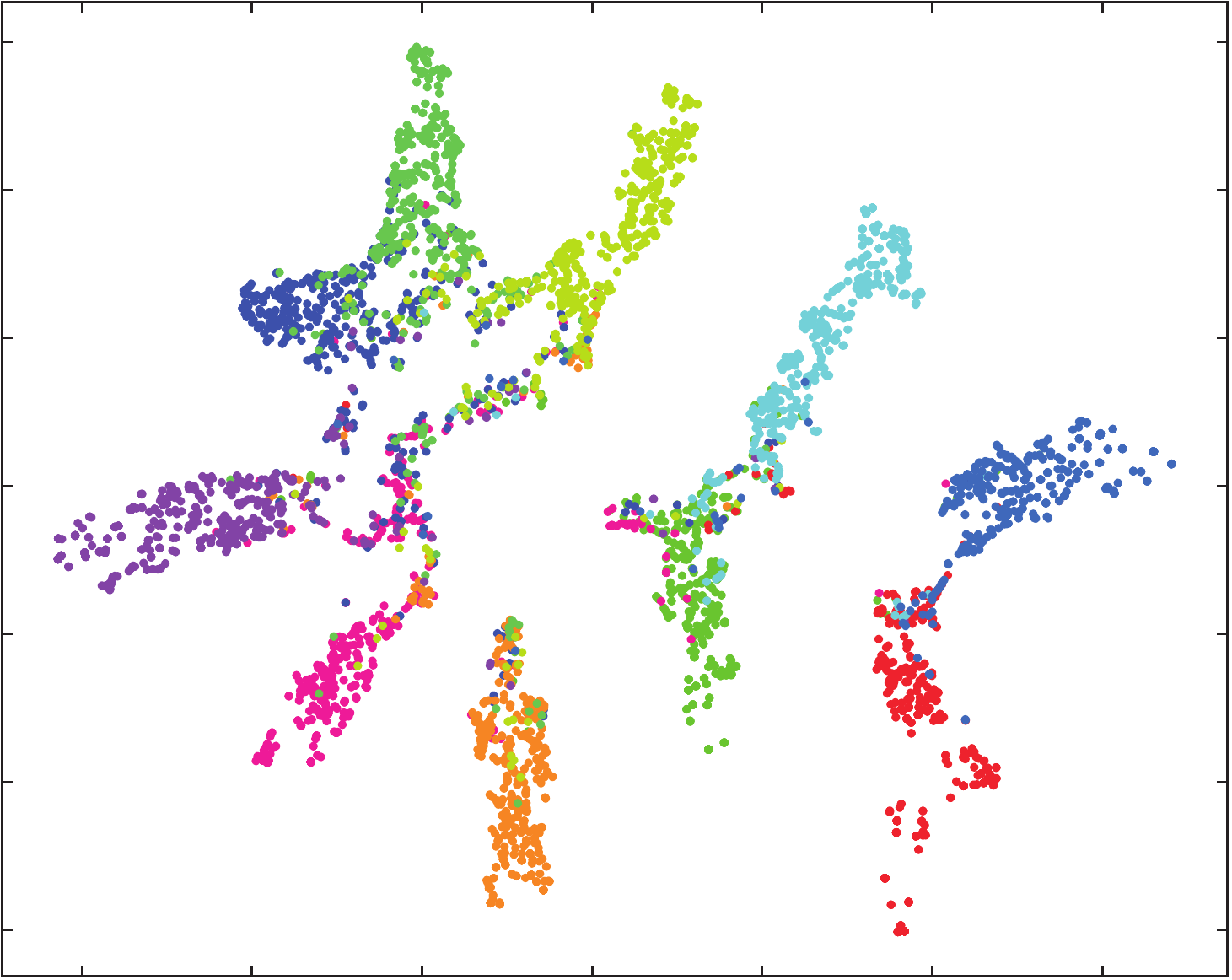}
			\caption{Ours}
			\vspace{-0.3cm}
		\end{subfigure}
		\caption{t-SNE visualization of 32 bits quantized features between DQN, DVSQ and DPQ. Images are randomly sampled on CIFAR-10's database, and samples with different labels are marked with different colors.}\label{fig.tSNE}
	\end{center}
\end{figure}

\subsection{Ablation Study} 
\label{AblationStudy}
In this subsection, we conduct a series of ablation studies on all three datasets to see the effect of each component. Specifically, we compare our method with the following changes: 1) $E$: which removes $L_S$ and $L_C$ from our model. The quantization is directly conducted on $fc7$ from pre-trained AlexNet; 2) $E + L_S$: which remove the $L_S$ loss; 3) $E + L_C$, which removes the $L_S$ loss; 4) {Two-Step}: which learns features and conducts quantization separately; 5) {No-Soft}: which removes $\ell_{quanH}$ and $\ell_{match}$ from Eq.~\ref{DistortionError}, and optimize hard assignments.
The results are shown in Tab.~\ref{tab.AblationStudy}.

From Tab.~\ref{tab.AblationStudy}, it can be observed that the best performance is achieved by using all the components of our model DPQ. Compared with $E$ which is an unsupervised model, our DPQ improves the mAP by 64.2\%, 64.5\%, 65.0\%, 62.6\% for CIFAR-10 dataset,  31.4\%, 20.1\%, 14.8\%, 12.1\% for NUS-WIDE dataset and 27.3\%, 33.7\%, 33.6\%, 34.0\% for ImageNet dataset. This indicates that the supervised information is very important for learning good representation. $E + L_S$ is a strong competitor, and its performance is very close to DPQ in NUS-WIDE dataset. However, for CIFAR-10 and ImageNet dataset, it is outperformed by DPQ with a large margin. This indicates that $L_C$ is very helpful to guide the learning of features. On the other hand, $E + L_C$, which also utilizes the supervision information by $L_C$, is not as good as $E + L_C$. This implies that $L_S$ is superior to $L_C$ for retrieval task. Unsurprisingly, DPQ outperforms Two-Step by large margin, especially in NUS-WIDE and ImageNet dataset. This indicates that in Two-Step method, suboptimal hash codes may be produced, since the quantization error is not statistically minimized and the feature representation is not optimally compatible with the quantization process. A joint learning of visual features and hash code can achieve better performance. Compared with No-Soft which only defines the loss function on the hard assignment, DPQ has significant performance gain. Soft-assignment plays the role of intermediate layer, and applying a loss function to soft-assignment is beneficial for hash codes learning.

\subsection{Qualitatively Results}

We randomly select 5,000 images from CIFAR-10 database and perform t-SNE visualization. Fig.~\ref{fig.tSNE} shows the results of DQN, DVSQ and DPQ. As a quantization method, the target is to cluster the data points with the same label, and separate the data points with different labels. Obviously, our DPQ performs the best compared with DQN and DVSQ. For both DQN and DVSQ, the data points with the same label may form several clusters which are far away. This may result in a low recall. On the other hand, the overlapping between different clusters in DQN and DVSQ is more serious than that of DPQ. This may cause a low retrieval precision of DQN and DVSQ, which is also reflected in Tab.~\ref{tab.Result}.

% ---------------------------------------------------------------------------------------
\section{Conclusion}
In this work, we propose a deep progressive quantization (DPQ) model, as an alternative to PQ, for large scale image retrieval. DPQ learns the quantization code sequentially, and approximates the original feature space progressively. Therefore, we can train the quantization codes with different code lengths simultaneously. Experimental results on the benchmark dataset show that our model significantly outperforms the state-of-the-art for image retrieval.

\section*{Acknowledgements}
This work is supported by the Fundamental Research Funds for the Central Universities (Grant No.~ZYGX2014J063, No.~ZYGX2016J085), the National Natural Science Foundation of China (Grant No.~61772116, No.~61872064, No.~61632007, No.~61602049).
%% The file named.bst is a bibliography style file for BibTeX 0.99c
{
\bibliographystyle{named}
\bibliography{ijcai19}
}
\end{document}